\documentclass[runningheads]{llncs}
\usepackage{graphicx}
\usepackage{amsmath,amssymb}
\usepackage{subfigure}
\begin{document}
\title{Selective Multi-Scale Learning for Object Detection}
%
%
\author{Junliang Chen\inst{1,2,3}\orcidID{0000-0001-7516-9546} \and
Weizeng Lu\inst{1,2,3}\orcidID{0000-0001-5215-6259} \and
Linlin Shen\inst{1,2,3}\orcidID{0000-0003-1420-0815}\thanks{Corresponding author}}

\authorrunning{J. Chen et al.}
%
\institute{Computer Vision Institute, School of Computer Science and Software Engineering, Shenzhen University, China \and Shenzhen Institute of Artificial Intelligence of Robotics of Society, Shenzhen, China \and Guangdong Key Laboratory of Intelligent Information Processing, Shenzhen University, Shenzhen 518060, China \\
\email{\{chenjunliang2016, luweizeng2018\}@email.szu.edu.cn}, \email{llshen@szu.edu.cn}
}
\maketitle              
\begin{abstract}
Pyramidal networks are standard methods for multi-scale object detection. Current researches on feature pyramid networks usually adopt layer connections to collect features from certain levels of the feature hierarchy, and do not consider the significant differences among them. We propose a better architecture of feature pyramid networks, named selective multi-scale learning (SMSL), to address this issue.
SMSL is efficient and general, which can be integrated in both single-stage and two-stage detectors to boost detection performance, with nearly no extra inference cost. RetinaNet combined with SMSL obtains 1.8\% improvement in AP (from 39.1\% to 40.9\%) on COCO dataset. When integrated with SMSL, two-stage detectors can get around 1.0\% improvement in AP.

\end{abstract}

\section{Introduction}
\label{sec:intro}

Object detection is a fundamental task in computer vision, whose target is to classify and locate all objects in an image. Image recognition aims to predict the probability of all classes for an image, and adopt the top probabilities and their corresponding classes as final result. Unlike image recognition where there is usually only one object in an image, in object detection, there usually exists various objects in the same image, with a wide range of scales. Therefore, it is difficult to represent different kinds of objects at the same feature representation level. To achieve this goal, a solution is to use multi-level feature representations. The features at higher levels are semantically strong with lower resolutions. While the low-level features are spatially finer with higher resolutions. Hence, the high-level features are more suitable for large-object detection while the low-level features are more beneficial for detecting smaller objects. 
The pyramidal architecture with multi-scale feature representations is widely used by many powerful object detectors \cite{Lin_2017_ICCV,NIPS2015_5638,tian2019fcos}.

One of the widely used pyramidal architecture is Feature Pyramid Networks (FPN) \cite{Lin_2017_CVPR}. FPN takes inputs from a backbone model, which is usually constructed for image recognition. The backbone model generates feature representations in different hierarchies with decreasing resolutions. FPN sequentially takes two adjacent layers from different levels in backbone as inputs, and combines them with a top-down pathway and lateral connections. The high-level features, with stronger semantic but lower resolution, are upsampled to fit the spatial size of the low-level features with higher resolution. Then a binary operation, usually element-wise summation, is conducted to aggregate the features. The low-level finer features are semantically enhanced after combination with high-level features from top-down pathway.

Although FPN is simple and effective for many detectors, there are some aspects to be improved. Path Aggregation Network (PANet) \cite{Liu_2018_CVPR} adds an additional bottom-up pathway on the base of FPN. This additional branch can strengthen the semantically enhanced features after FPN, with finer spatial features at lower levels. Balanced Feature Pyramid (BFP) \cite{Pang_2019_CVPR} gathers cross-level features from FPN or other pyramidal architectures to the same level. Then a refinement module is carried out after element-wise average of the gathered features. The averaged features, a fusion of features cross all levels, can be considered as global information. The fused features are then scattered to all levels and summed up with the original input features. This process merges the original features with features from all other levels, enabling detectors to perceive information from all levels. Recent researches \cite{Kong_2018_ECCV,Kim_2018_ECCV} explore better connections of cross-scale features to produce a pyramidal architecture for feature representations.

However, the above works ignore the variances among features from different scales and give them the same weights for combination, or only merge features from partial scales. Inspired by these, we propose an architecture, named selective multi-scale learning (SMSL), to dynamically learn a better feature representation for each level from multi-scale features. SMSL can efficiently improve the detection performance of both single-stage and two-stage detectors with only a small increase of inference cost.


In this study, we make the following contributions:
\begin{itemize}
    \item We propose the selective multi-scale learning (SMSL) to generate specific features for each level by selectively merge features from multi scales.
    \item Combined with SMSL, RetinaNet achieves performance of 44.3\% AP on COCO dataset.
    \item The proposed framework can also be applied to two-stage object detectors to improve the detection performance.
\end{itemize}

\section{Related Work}

Recognizing multi-scale objects is a fundamental but challenging task in computer vision. Pyramidal feature representations is a general technique \cite{adelson1984pyramid} in this area. A simple method is to use convolutional networks (ConvNets) to extract features from image pyramids. However, this method brings huge computation burden, as the ConvNets forward repeatedly for the same image. To solve the problem,  an effective solution is to directly take advantage of the features generated by the ConvNets, instead of using image pyramids. Recent researches \cite{Lin_2017_ICCV,He_2017_ICCV,Cai_2018_CVPR,newell2016stacked} propose many cross-scale connections to connect multi-level features from the ConvNets. Though keep the original resolution, the connected features are semantically enhanced.

\textbf{Partial connections.} Partial connections are one of the standard pyramidal architectures. FPN \cite{Lin_2017_CVPR} connects two adjacent layers in the top-down pathway by upsampling the high-level features to fit the size of the features at lower level and element-wisely sum up them. This architecture enhances the low-level features with stronger semantic information from higher levels. Although FPN is simple and effective to improve feature representations, the features still lack information from lower levels. To address this problem, Liu et.al \cite{Liu_2018_CVPR} propose Path Aggregation Network (PANet) to add an accessional bottom-up pathway on the basis of FPN. In PANet, low-level features are downsampled and summed up with features at higher level. Therefore, the semantically enhanced features after FPN can obtain finer spatial information. NAS-FPN \cite{Ghiasi_2019_CVPR} uses Neural Architecture Search (NAS) algorithm to discover a better pyramidal architecture covering all cross-scale connections.

\textbf{Full connections.} Another way to integrate multi-level features is to gather features and fuse them to generate features for different levels and scatter the fused features to the corresponding level. Kong et.al \cite{Kong_2018_ECCV} first gather multi-level features and combine them, then use global attention for further refinement. After that, the local reconfiguration module is employed to further capture local information. The produced features are resized and element-wisely summed up with the original input which is linearly projected by a $1 \times 1$ convolution. Balanced Feature Pyramid (BFP) \cite{Pang_2019_CVPR} gathers features to a level and applies element-wise averaging. Then a non-local module is utilized to refine the integrated features, which are then scattered to all levels. The refined features are element-wisely summed up with the original input features at each level.


The above methods obtain features from partial scales, and usually merge them through linear operation (such as element-wise summation) which gives features from different scales the same weights. However, for a specific level, the features from different scales have different importance. Therefore, the detector should learn to selectively merge the multi-scale features.

\section{Selective Multi-Scale Learning}

\subsection{Network Architecture}

\begin{figure}
    \centering
    \subfigure[]{
        \includegraphics[width=0.46\textwidth]{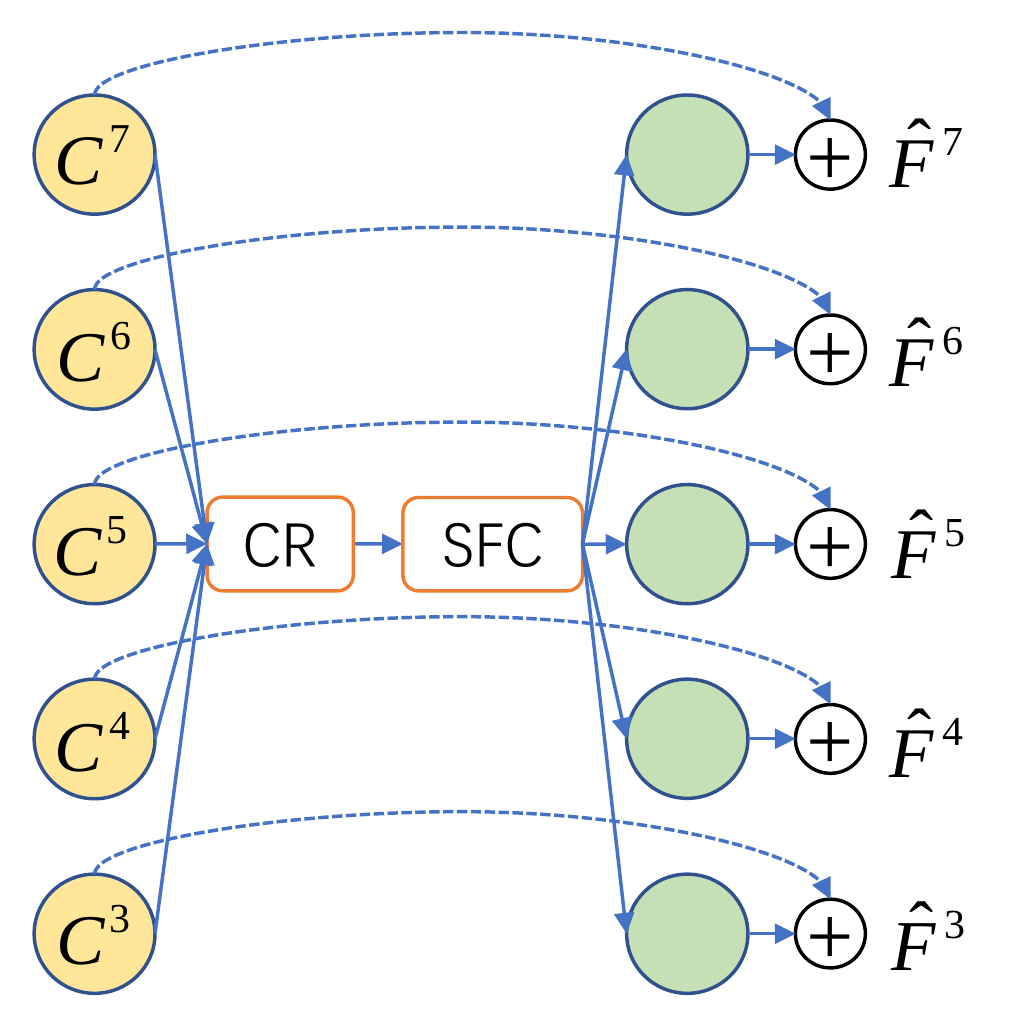}
    } \quad
    \subfigure[]{
        \includegraphics[width=0.35\textwidth]{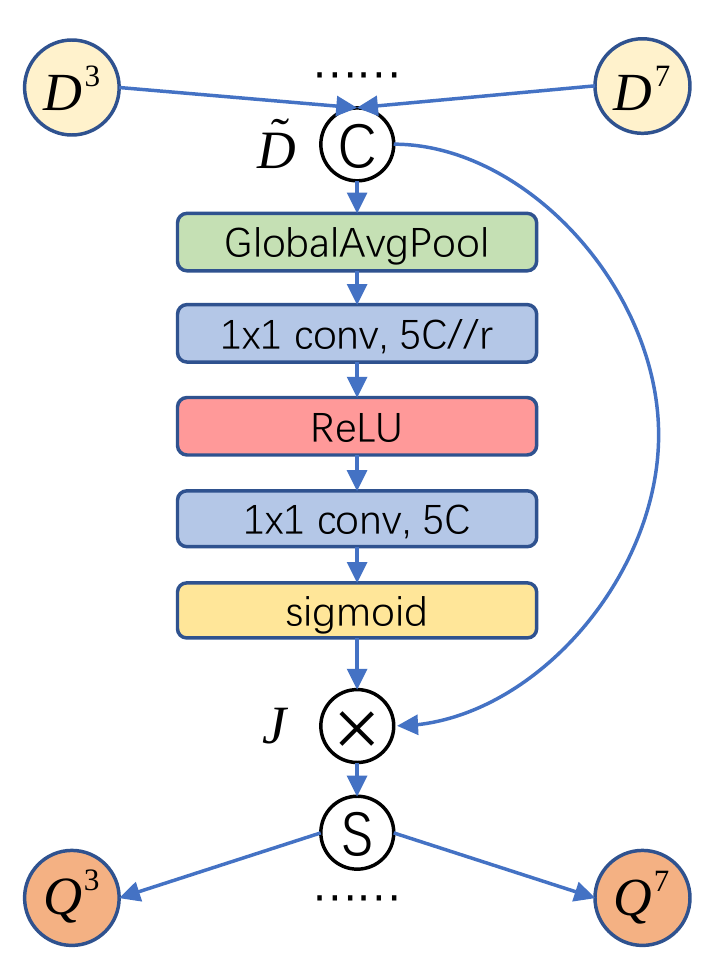}   
    }
    \caption{(a) The overview of the proposed selective multi-scale learning. (b) The channel rescaling module. ``\textcircled{C}" and ``\textcircled{S}" denote channel concatenation and channel spliting, respectively.}
    \label{fig:smsl_cr}
\end{figure}

\subsubsection{Overview}

\begin{figure}
    \centering
    \subfigure[]{
        \includegraphics[width=0.48\textwidth]{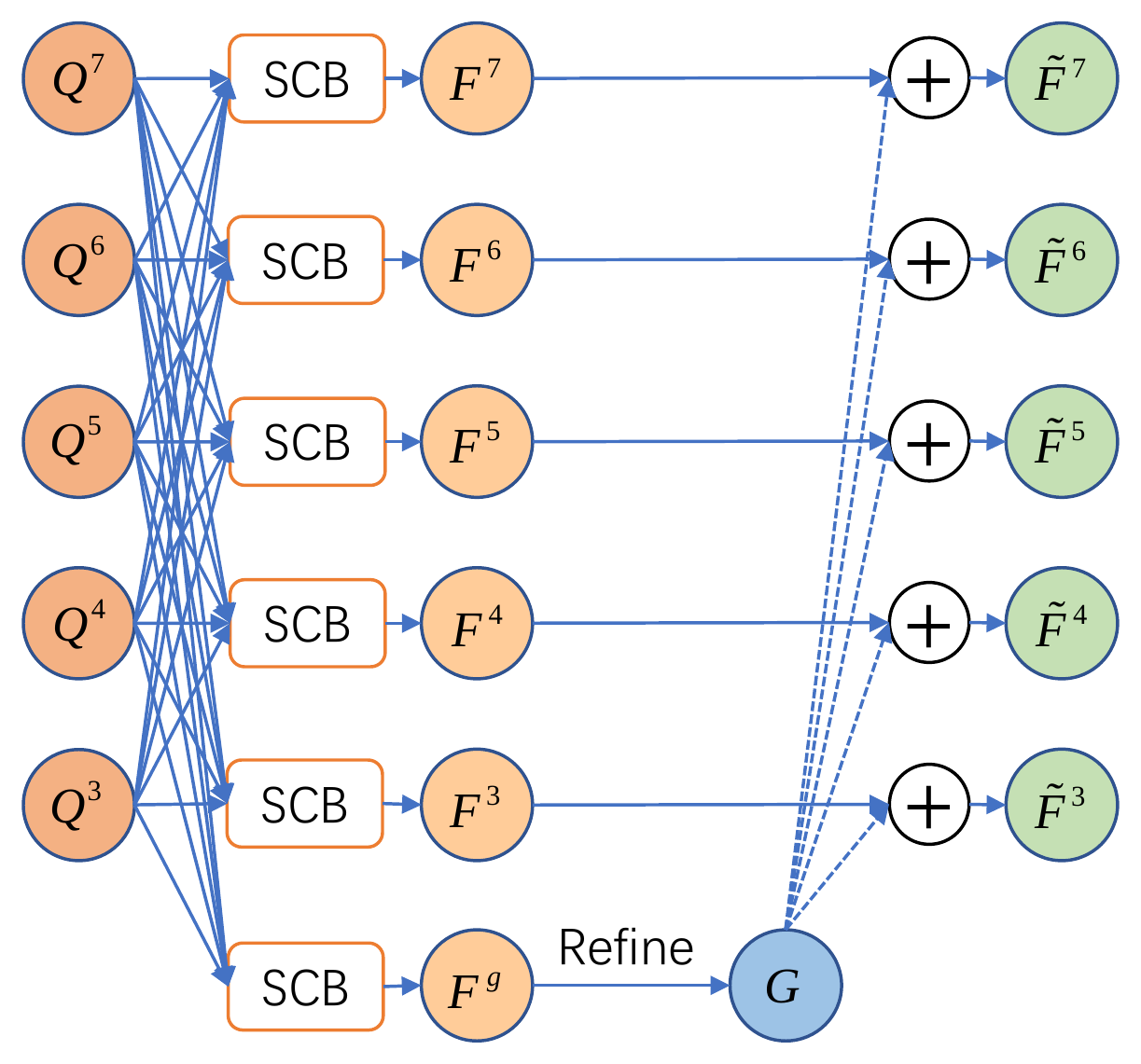}
    } \quad
    \subfigure[]{
        \includegraphics[width=0.36\textwidth]{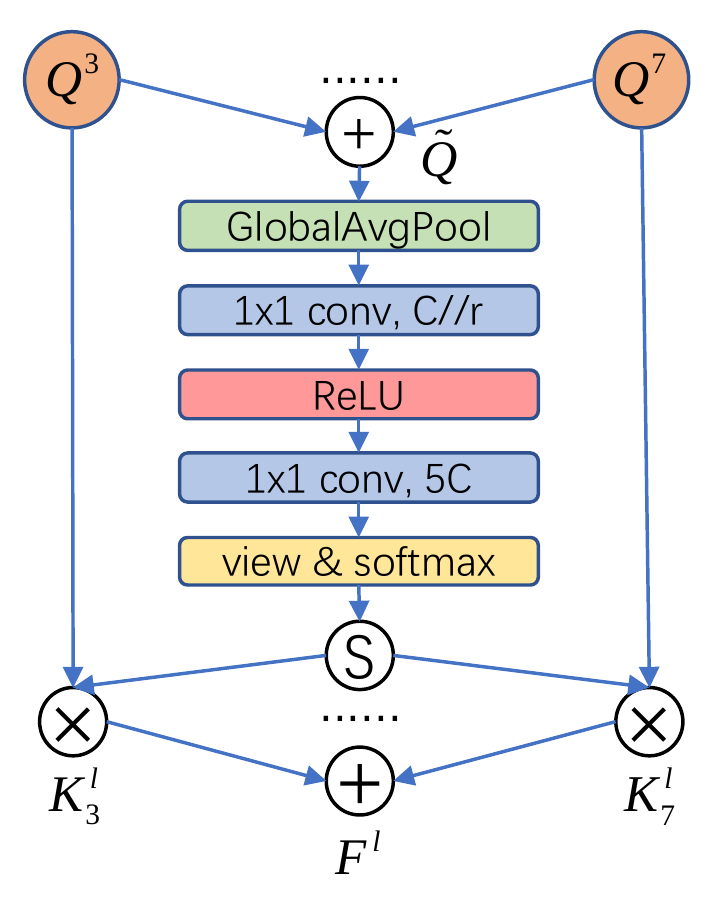}
    }
    \caption{(a) The selective feature combination module. ``SCB" deontes the selective combination module. (b) An selective combination module for the $l$-th level. ``\textcircled{S}" denotes channel spliting.}
    \label{fig:cr}
\end{figure}

Fig \ref{fig:smsl_cr}(a) shows the architecture of selective multi-scale learning. We use the $\left\{C^{3}, C^{4}, C^{5}\right\}$ layers from ResNet \cite{He_2016_CVPR} backbone. Then we generate $C^{6}$ and $C^{7}$ layers by separately applying a $3\times3$ convolution with stride 2 on $C^{5}$ and $C^{6}$ layers. Therefore, the original inputs are $\left\{C^{3}, C^{4}, C^{5}, C^{6}, C^{7}\right\}$, which are gathered to a level and then passed to channel rescaling (CR) module (Fig \ref{fig:smsl_cr}(b)) and selective feature combination (SFC) module (Fig \ref{fig:cr}(a)) to generate level-specific features. At each level, the generated features are then element-wisely summed up with the corresponding input as the final output.


\subsubsection{Channel Rescaling.}
\label{sec:integration}

The features at level $l$ after resizing are denoted as $\mathbf{D}^l \in \mathbb{R}^{C\times H \times W}$ with a resolution $H \times W$, and the indexes of the input levels with lowest and highest resolution are denoted as $l_{min}$ and $l_{max}$. Let $L$ be the number of levels, then $L=l_{max}-l_{min}+1$. In our experiments, the gather level is set to $(l_{min} + l_{max})/2$.

The context of the multi-level features at each channel is different, so the importance of features at each channel is as well different. Therefore, we aim to emphasize the important features and suppress the less useful features, which can be regarded to select the information via a gate. To achieve this goal, we propose channel rescaling module to rescale the features at different channels of the multi-level features. After gathering the multi-level features, we first concat them as:
\begin{equation}
    \widetilde{\mathbf{D}}=\left[ \mathbf{D}^{l_{min}}, \dots, \mathbf{D}^l, \dots, \mathbf{D}^{l_{max}} \right]
\end{equation} where $\widetilde{\mathbf{D}} \in \mathbb{R}^{LC \times H \times W}$. Then channel rescaling (CR) is accomplished by the following steps.

For a specific channel of $\widetilde{\mathbf{D}}$, we get the global information by using global average pooling (GAP). We denote the result after GAP as $\mathbf{x}$. The result of the $c'$-th channel can be calculated as:
\begin{equation}
    x_{c'}=\frac{1}{HW}\sum_{i=1}^{H}\sum_{j=1}^{W} \widetilde{D}_{c', i, j}
\end{equation}

We then generate the weights for each channel by two fully connected (FC) layers followed by the sigmoid function:
\begin{equation}
    \mathbf{s}=\sigma(\mathbf{W_2}\delta(\mathbf{W_1}\mathbf{x}))
\end{equation} where $\mathbf{W_1} \in \mathbb{R}^{\frac{LC}{r} \times LC}$, $\mathbf{W_2} \in \mathbb{R}^{LC \times \frac{LC}{r}}$, $\delta$ is the ReLU function and $\sigma$ denotes the sigmoid function. $r$ is the reduction ratio, and is set to 8 in our experiments.

We denote the output after channel resclaing module as $\mathbf{J}$. For the $c'$-th channel, the output $\mathbf{J}_{c'}$ is generated by rescaling $\widetilde{\mathbf{D}}_{c'}$ with $s_{c'}$:
\begin{equation}
    \mathbf{J}_{c'}=s_{c'} \otimes \widetilde{\mathbf{D}}_{c'}
\end{equation} where $\otimes$ denotes the channel-wise multiplication.

Then we split $\mathbf{J}$ into $L$ groups:
\begin{equation}
    \mathbf{J}=\left[ \mathbf{Q}^{l_{min}}, \dots, \mathbf{Q}^l, \dots, \mathbf{Q}^{l_{max}} \right]
\end{equation}
\begin{equation}
    \mathbf{Q}^l=\mathbf{J}_{1 + (l - 1)C:lC, :, :}
\end{equation} where $l \in \left\{l_{min}, \dots, l_{max}\right\}$.

\subsubsection{Selective Feature Combination}
\ \\
\textbf{Local Feature.} A simple approach is to scatter the $L$ rescaled features $\mathbf{Q}=\left\{\mathbf{Q}^{l_{min}}, \dots, \mathbf{Q}^{l_{max}}\right\}$ to all levels. However, as mentioned before, features at different levels have various semantic contexts and are thus suitable to detect objects with different sizes. In addition, the features scattered to $l$-th level shall put more emphasis on the neighboring levels, i.e. $\left\{\mathbf{Q}^{l-1}, \mathbf{Q}^l,  \mathbf{Q}^{l+1}\right\}$, as they have more similar semantic contexts. 

Motivated by these observations, we design a selective combination (SFC) module (Fig \ref{fig:cr}(b)) to combine the set of $\mathbf{Q}$ features to generate local feature $\mathbf{F}^l$, which is to be scattered to the $l$-th level. The $C$ channel feature $\mathbf{F}^l=\{\mathbf{F}^l_1, \dots, \mathbf{F}^l_c, \dots, \mathbf{F}^l_C\}$ is a weighted combination of $\mathbf{Q}$. The weights are different for each target level and learned by the following steps.

Our goal is to adaptively select features from different levels. An effective idea is to use gate to control the information flow from multiple levels. To achieve this goal, we should aggregate the features from multiple levels. A simple way is to use concatenation to merge the features, but this requires more parameters. Therefore, we use element-wise summation to merge features from multiple levels:
\begin{equation}
    \mathbf{\widetilde{Q}}=\sum_{i=l_{min}}^{l_{\max }} \mathbf{Q}^{i}
\end{equation} then we create the global context $\mathbf{g} \in \mathbb{R}^{C}$ by simply using average pooling, the $c$-th element of the global context can be formulated as:
\begin{equation}
    g_c=\frac{1}{H \times W}\sum_{i=1}^{H}\sum_{j=1}^{W} \widetilde{Q}_{c, i, j}
\end{equation}

Next, we compact the global context into feature $\mathbf{z}^l \in \mathbb{R}^{\frac{C}{r}}$ to guide the adaptive selection. $r$ is the reduction ratio, and is set as 8 in our experiments. To achieve this, we apply a fully connected (FC) layer to generate the result:
\begin{equation}
    \mathbf{z}^l=\delta(\mathcal{L}(\mathcal{FC}^l_1(\mathbf{g})))=\delta(\mathcal{L}(\mathbf{W}^l \mathbf{g}))
\end{equation} where $\mathbf{W}^l \in \mathbb{R}^{\frac{C}{r} \times C}$, $\mathcal{L}$ denotes the Layer Normalization \cite{Ba2016LayerN} and $\delta$ is the ReLU function \cite{Nair2010RectifiedLU}.

To adaptively select features from different levels, a soft attention across channels is employed. The soft attention is a channel-wise weight generated under the guidance of the compacted global context $\mathbf{z}^l$. We first generate the original weight $\mathbf{U}^l \in \mathbb{R}^{LC}$ using an FC layer:
\begin{equation}
    \mathbf{U}^l=\mathcal{FC}^l_2(\mathbf{z}^l)=\mathbf{V}^l \mathbf{z}^l
\end{equation} where $\mathbf{V}^l \in \mathbb{R}^{C \times LC}$. Then we reshape the weight $\mathbf{U}^l \in \mathbb{R}^{LC}$ into $\mathbf{M}^l \in \mathbb{R}^{L \times C}$. Let $\mathbf{A}^l \in \mathbb{R}^{L \times C}$ be the soft attention weight for $\left\{\mathbf{Q}^{l_{min}}, \dots, \mathbf{Q}^{l_{max}}\right\}$.

For a specific level $i$ and channel $c$, the soft attention weight $A^l_{i, c}$ can be computed as:

\begin{equation}
    A^l_{i, c}=\frac{e^{\mathbf{M}^l_{i, c}}}{\sum_{j=l_{min}}^{l_{max}}e^{\mathbf{M}^l_{j, c}}}
\end{equation}

After adaptive selection, all the features from different levels own their specific weights in the channel-wise aspect. The final output $\mathbf{F}^l \in \mathbb{R}^C$ are the weighted summation of multi-level features via the soft attention weights. For the $c$-th channel, the output $\mathbf{F}^l_{c}$ can be calculated as:

\begin{equation}
    \mathbf{K}^l_{i, c} = A^l_{i, c} \otimes \mathbf{Q}^i_c,\ i \in \{l_{min}, \dots, l_{max}\}
\end{equation}

\begin{equation}
    \mathbf{F}^l_{c}=\sum_{i=l_{min}}^{l_{max}} \mathbf{K}^l_{i, c}
\end{equation} where $\otimes$ denotes the channel-wise multiplication.


\textbf{Global Feature.} As global context has been widely used in rescaling and weight features of different levels, we argue that a global feature represents the overall information of all levels shall also be learned and injected into the $L$ local features $\mathbf{F}^l$, before they are scattered to the target levels. The same combination process described above can be used to learn the weights of $\mathbf{Q}$, which can be used to calculate the global feature $\mathbf{F}^g$. We use the non-local \cite{Wang_2018_CVPR} module with embed Gaussian attention to further refine $\mathbf{F}^g$ to $\mathbf{G}$. As justified by the ablation study in experimental section, the inclusion of $\mathbf{G}$ can further increase the performance of the feature pyramid network.

The $L$ local features to be scattered to the $l$-th target level, $\widetilde{\mathbf{F}}^l$, can now be calculated as the element-wise summation of $\mathbf{F}^l$ and global feature $\mathbf{G}$:
\begin{equation}
    \widetilde{\mathbf{F}}^l=\mathbf{F}^l \oplus \mathbf{G}
\end{equation} where $\oplus$ denotes the element-wise summation.

After feature fusion, the fused features are then scattered to the same size as the input of the corresponding level via resizing. For the $l$-th level, the final features $\hat{\mathbf{F}}^l$ can be computed as:
\begin{equation}
    \hat{\mathbf{F}}^l=Resize(\widetilde{\mathbf{F}}^l) \oplus \mathbf{C}^l
\end{equation} where \textit{Resize} denotes the resizing function, $\oplus$ denotes the element-wise summation.

\section{Experiments}

\subsection{Dataset and Evaluation Metrics}
We conduct our experiments on the COCO dataset \cite{10.1007/978-3-319-10602-1_48}. For training, we use the data in \textit{train-2017} split, which contains 115k images. For ablation study, we use the data in the \textit{val-2017} split consisting of 5k images as validation. We report our main results on the \textit{test-dev} (20k images without public annotations available) split. All the results are reported in the standard COCO-style Average Precision ($AP$) metrics.

\subsection{Implementation Details}
For fair comparisons, all the experiments are conducted on the MMDetection \cite{mmdetection} platform. If not specified, for all other hyper-parameters, we follow the same settings in MMDetection \cite{mmdetection} for fair comparison.

\subsubsection{Training details.} The training settings are as follows if not specified. We use ResNet-50 \cite{He_2016_CVPR} as our backbone networks, and RetinaNet \cite{Lin_2017_ICCV} as our detector. The backbone network is initialized with the pretrained model on ImageNet \cite{5206848}. We use the stochastic gradient descent (SGD) optimizer to train our networks for 12 epochs with batch size 16. The initial learning rate is 0.01 and divided by 10 after 8 and 11 epochs. The input images are resized to have a resolution of $\sim 1333 \times 800$.


\begin{table}[b]
\centering
\renewcommand\arraystretch{1.2}
\renewcommand\tabcolsep{2mm}
\caption{Ablation studies on component effectiveness on COCO \textit{val-2017}, with ResNet-50 \cite{He_2016_CVPR} backbone. ``LF", ``GF", and ``CR" denote Local Features, Global Feature, and Channel Rescaling respectively.}
\begin{tabular}{cccccccccc}
\hline
LF & GF & CR & $AP$ & $AP_{50}$ & $AP_{75}$ & $AP_S$ & $AP_M$ & $AP_L$ \\ \hline
\checkmark &            &            & 35.5 & 55.5 & 37.7 & 20.8 & 39.7 & 46.2 \\ 
\checkmark &            & \checkmark & 35.8 & 56.5 & 37.9 & 21.1 & 40.2 & 46.4 \\ 
\checkmark & \checkmark &            & 35.8 & 56.0 & 38.0 & 20.5 & 40.0 & 47.1 \\ 
\checkmark & \checkmark & \checkmark & \textbf{36.1} & \textbf{56.6} & \textbf{38.4} & \textbf{21.2} & \textbf{40.3} & \textbf{47.3} \\
\hline \\
\end{tabular}
\label{table:ablation}
\end{table}

\begin{table}[t]
\centering
\renewcommand\arraystretch{1.2}
\renewcommand\tabcolsep{2mm}
\caption{Application in other pyramidal architectures based on RetinaNet detector (1st group) and two-stage detectors (2nd group) on COCO \textit{val-2017}. ``*" denotes our re-implementation. ``Params" denotes the number of total parameters (M) and ``Time" denotes the inference time (ms) on single Tesla P100.}
\begin{tabular}{lcccccc}
\hline
 & SMSL & $AP$ & $AP_{50}$ & $AP_{75}$ & Params (M) & Time (ms)\\
\hline
Arch \\ 
FPN & & 35.5 & 55.3 & 37.9 & 37.74 & 96.8\\ 
FPN & \checkmark & 36.4[\textbf{+0.9}] & 56.9 & 38.9 & 38.72 & 99.0\\ 
\hline
PANet* & & 35.9 & 55.8 & 38.4 & 39.51 & 100.6\\ 
PANet* & \checkmark & 37.0[\textbf{+1.1}] & 57.6 & 39.4 & 40.49 & 101.6\\ 
\hline
\hline
Detector \\
Mask & & 35.2 & 56.4 & 37.9 & 44.18 & 92.6\\ 
Mask & \checkmark & 36.0[\textbf{+0.8}] & 57.6 & 38.6 & 45.15 & 97.2\\ 
\hline
Cascade & & 38.1 & 55.9 & 41.1 & 69.17 & 84.0\\ 
Cascade & \checkmark & 39.1[\textbf{+1.0}] & 57.5 & 42.2 & 70.15 & 88.0\\ 
\hline \\
\end{tabular}
\label{table:application}
\end{table}

\begin{table*}[t]
\centering
\renewcommand\arraystretch{1.4}
\renewcommand\tabcolsep{2mm}
\caption{Comparisons with mainstream methods on COCO \textit{test-dev}. ``\dag" denotes results under multi-scale testing.}
\scriptsize{
\begin{tabular}{l|l|cccccc}
\hline
Method & Backbone & $AP$ & $AP_{50}$ & $AP_{75}$ & $AP_{S}$ & $AP_{M}$ & $AP_{L}$ \\ \hline
\textit{Two-stage methods} & & \\
Faster R-CNN \cite{Lin_2017_CVPR} & ResNet-101 &36.2 & 59.1 & 39.0 & 18.2 & 39.0 & 48.2 \\ 
Mask R-CNN \cite{He_2017_ICCV} & ResNeXt-101 & 39.8 & 62.3 & 43.4 & 22.1 & 43.2 & 51.2 \\ 
LH R-CNN \cite{DBLP:journals/corr/abs-1711-07264} & ResNet-101 & 41.5 & - & - & 25.2 & 45.3 & 53.1 \\ 
Cascade R-CNN \cite{Cai_2018_CVPR} & ResNet-101 & 42.8 & 62.1 & 46.3 & 23.7 & 45.5 & 55.2 \\ 
TridentNet \cite{li2019scale} & ResNet-101-DCN & 48.4 & 69.7 & 53.5 & 31.8 & 51.3 & 60.3 \\ 
\hline
\textit{Single-stage methods} & & \\ 
ExtremeNet \cite{Zhou_2019_CVPR} & Hourglass-104 & 40.2 & 55.5 & 43.2 & 20.4 & 43.2 & 53.1 \\ 
FoveaBox \cite{kong2019foveabox} & ResNet-101 & 40.6 & 60.1 & 43.5 & 23.3 & 45.2 & 54.5 \\ 
FoveaBox \cite{kong2019foveabox} & ResNeXt-101 & 42.1 & 61.9 & 45.2 & 24.9 & 46.8 & 55.6 \\ 
CornerNet \cite{Law_2018_ECCV} & Hourglass-104 & 40.5 & 56.5 & 43.1 & 19.4 & 42.7 & 53.9 \\ 
CornerNet \cite{Law_2018_ECCV}$^{\dag}$ & Hourglass-104 & 42.2 & 57.8 & 45.2 & 20.7 & 44.8 & 56.6 \\ 
FreeAnchor \cite{zhang2019freeanchor} & ResNet-101 & 43.1 & 62.2 & 46.4 & 24.5 & 46.1 & 54.8 \\
FreeAnchor \cite{zhang2019freeanchor} & ResNeXt-101 & 44.9 & 64.3 & 48.5 & 26.8 & 48.3 & 55.9 \\ 
FSAF \cite{zhu2019feature} & ResNet-101 & 40.9 & 61.5 & 44.0 & 24.0 & 44.2 & 51.3 \\ 
FSAF \cite{zhu2019feature} & ResNeXt-101 & 42.9 & 63.8 & 46.3 & 26.6 & 46.2 & 52.7 \\ 
FCOS \cite{tian2019fcos} & ResNet-101 & 41.5 & 60.7 & 45.0 & 24.4 & 44.8 & 51.6 \\
FCOS \cite{tian2019fcos} & ResNeXt-101 & 44.7 & 64.1 & 48.4 & 27.6 & 47.5 & 55.6 \\ 
ATSS \cite{zhang2020bridging} & ResNet-101 & 43.6 & 62.1 & 47.4 & 26.1 & 47.0 & 53.6 \\ 
Dense RepPoints \cite{yang2019dense} & ResNeXt-101-DCN & 48.9 & 69.2 & 53.4 & 30.5 & 51.9 & 61.2 \\ 
\hline
\hline 
RetinaNet \cite{Lin_2017_ICCV} & ResNet-101 & 39.1 & 59.1 & 42.3 & 21.8 & 42.7 & 50.2 \\ 
RetinaNet (ours) & ResNet-101 & 40.9 & 62.3 & 44.1 & 25.1 & 44.7 & 49.9 \\ 
RetinaNet (ours)$^{\dag}$ & ResNet-101 & 42.7 & 63.8 & 46.3 & 27.8 & 45.1 & 52.5 \\ 
\hline 
RetinaNet \cite{Lin_2017_ICCV} & ResNeXt-101 & 40.8 & 61.1 & 44.1 & 24.1 & 44.2 & 51.2 \\ 
RetinaNet (ours) & ResNeXt-101 & 42.6 & 64.4 & 45.7 & 26.7 & 46.3 & 51.8 \\ 
RetinaNet (ours)$^{\dag}$ & ResNeXt-101 & 44.3 & 65.5 & 48.2 & 29.4 & 46.9 & 54.5 \\ 
\hline 
\end{tabular}}
\label{table:comp_sota}
\end{table*}

\subsubsection{Inference details.} The inference settings are as follows if not specified. For inference, we first select the top 1000 confidence predictions from each prediction layer. Then, we use a confidence threshold of 0.05 to filter out the predictions with low confidence for each class. Then, we apply non-maximum suppression (NMS) to the filtered predictions for each class separately with a threshold of 0.5. Finally, we adopt the predictions with top 100 confidences for each image as the final results.

\subsection{Ablation Study}

As our selective multi-scale learning approach mainly consists of two steps, i.e. CR and SFC, we firstly justify the importance of the proposed module using ablation study. As the local features (LF) are necessary to scatter to the feature pyramid, we only perform an ablation study on the global feature (GF) included in the combination module. The two modules, i.e. CR and GF are removed to see their effects on the performance of the baseline, which are shown in Table \ref{table:ablation}. 

The second row in the table suggests that the CR module improve the overall AP of baseline from 35.5\% to 35.8\%. 
Compared to the baseline, the adoption of GF improves $AP$ and $AP_{50}$ by 0.3\% and 1.1\%, respectively. When both modules are used, the $AP$ is further improved to 36.1\%. In summary, both CR and GF can enhance the features and effectively boost the detection performance, which justify the usefulness of our approach.

\subsection{Application in Pyramid Architectures.}

In this section, we evaluate the effectiveness of our method on different pyramidal architectures by combining them with our method. As shown in the 1st group of Table \ref{table:application}, when combined with SMSL, FPN \cite{Lin_2017_CVPR} and PANet \cite{Pang_2019_CVPR} get 0.9\% and 1.1\% improvement in AP respectively, with only a small increase of parameters and little extra inference time (+2.2 ms and +1.0 ms, respectively).


\subsection{Application in Two-Stage Detectors}

In this section, we conduct experiments to evaluate the effectiveness of our method on two-stage detectors, including Mask R-CNN \cite{He_2017_ICCV} and Cascade R-CNN \cite{Cai_2018_CVPR}. The resolution of the input image is set to $640 \times 640$. The batch size is adjusted according to the memory limitation with a linearly scaled learning rate. As shown in the 2nd group of Table \ref{table:application}, when combined with SMSL, Mask R-CNN and Cascade R-CNN get 0.8\% and 1.0\% improvement in AP, with nearly no extra inference time (+4.6 ms and +4.0 ms, respectively). The results justify the effectiveness of our method on two-stage detectors.

\subsection{Comparisons with Mainstream Methods}

After ablation study and comparison with pyramidal networks, we now compare our approach with mainstream methods in Table \ref{table:comp_sota}. Both single-stage and two-stage detectors are included for comparison. We report the performance of our SFPN using both ResNet-101 and ResNeXt-101 backbones. We adopt $2 \times$ longer training with scale-jitter. For ResNeXt-101 backbone, due to memory limitation, we train the detector using batch size 12 with a linearly scaled learning rate. 

As shown in Table \ref{table:comp_sota}, combined with our method, RetinaNet with ResNet-101 backbone get 1.8\% improvement in AP. With ResNeXt-101 backbone and single-scale setting, RetinaNet with our method achieves 42.6\% AP, which is close to two-stage detectors, such as Cascade R-CNN (42.8\% AP). If multi-scale test is adopted, the best performance of RetinaNet can be further boosted to 44.3\% AP, which surpasses many mainstream object detectors.

\section{Conclusions}
In this paper, we propose selective multi-scale learning, which considers the different importance of the cross-scale features and selectively combine multi-scale features. SMSL can effectively improve the detection performance of single-stage detector, with almost no extra inference cost. The experimental results shows that SMSL can also be applied to two-stage detectors to boost the detection performance.

\section{Acknowledgments} This work was supported by National Natural Science Foundation of China under Grant 91959108.

%
%
%
\bibliographystyle{splncs04}
\bibliography{references}
%




\end{document}